\documentclass[sigconf]{aamas} 
\usepackage{balance} 

\usepackage[utf8]{inputenc}

\usepackage{amsmath}
\usepackage{amsthm}
\usepackage{amsfonts}
\usepackage{algpseudocode}
\usepackage{algorithm}

\newtheorem{theorem}{Theorem}
\newtheorem{lemma}{Lemma}
\newtheorem{corol}{Corollary}
\newtheorem{asumption}{Asumption}

\newcommand{\R}{\mathbb{R}}

\newcommand{\D}{\mathbb{D}}

\usepackage{todonotes} 

\title{Decentralized model-free reinforcement learning in stochastic games with average-reward objective}
\date{October 2022}

\setcopyright{ifaamas}
\acmConference[AAMAS '23]{Proc.\@ of the 22nd International Conference
on Autonomous Agents and Multiagent Systems (AAMAS 2023)}{May 29 -- June 2, 2023}
{London, United Kingdom}{A.~Ricci, W.~Yeoh, N.~Agmon, B.~An (eds.)}
\copyrightyear{2023}
\acmYear{2023}
\acmDOI{}
\acmPrice{}
\acmISBN{}

\setcopyright{none}
\renewcommand\footnotetextcopyrightpermission[1]{} 
\author{Romain Cravic}
\affiliation{
  \institution{Univ. Grenoble Alpes, Inria, CNRS, Grenoble INP, LIG}
  \city{38000 Grenoble}
  \country{France}}
\email{romain.cravic@inria.fr} 
\author{Nicolas Gast}
\affiliation{
  \institution{Univ. Grenoble Alpes, Inria, CNRS, Grenoble INP, LIG}
  \city{38000 Grenoble}
  \country{France}}
\email{nicolas.gast@inria.fr}

\author{Bruno Gaujal}
\affiliation{
  \institution{Univ. Grenoble Alpes, Inria, CNRS, Grenoble INP, LIG}
  \city{38000 Grenoble}
  \country{France}}
\email{bruno.gaujal@inria.fr}

\begin{document}

\begin{abstract}
    We propose the first model-free algorithm that achieves low regret performance for decentralized learning in two-player zero-sum tabular stochastic games with infinite-horizon average-reward objective. In decentralized learning, the learning agent controls only one player and tries to achieve low regret performances against an arbitrary opponent. This contrasts with centralized learning where the agent tries to approximate the Nash equilibrium by controlling both players. In our infinite-horizon undiscounted setting, additional structure assumptions is needed to provide good behaviors of learning processes : here we assume for every strategy of the opponent, the agent has a way to go from any state to any other. This assumption is the analogous to the "communicating" assumption in the MDP setting. We show that our Decentralized Optimistic Nash Q-Learning (DONQ-learning) algorithm achieves both sublinear high probability regret of order $T^{3/4}$ and sublinear expected regret of order $T^{2/3}$. Moreover, our algorithm enjoys a low computational complexity and low memory space requirement compared to the previous works of \cite{wei2017online} and \cite{jafarnia2021learning} in the same setting.
\end{abstract}

\maketitle

\section{Introduction}

Recently, reinforcement learning, combined with technological\linebreak progress, has achieved superhuman performances in computing optimal behaviors in various decision processes. Particularly, there is a development of self-play algorithms to design strong artificial intelligence in games such as Go or Chess \cite{silver2017mastering,silver2018general}. These methods have the practical interest that they do not require supervision neither with huge data set nor with expert feedback to train efficiently. We focus here on Competitive Multi-Agent Reinforcement Learning (MARL) where several agents interact with an environment that may involve randomness, in order to maximize their own profit. One of the mostly used model to deal with MARL is stochastic game (a.k.a. Markov game in the literature) that was introduced in the 50s by Shapley \cite{shapley1953stochastic}. Stochastic games (SGs) are more pertinent to handle MARL compared to Markov decision processes (MDPs) that are widely used for single-agent reinforcement learning. 

Here we consider learning in infinite-horizon two-player zero-sum game with the average-reward criterion: There are two players that play indefinitely, and each player wins exactly what their opponent loses. At each time, the game has a current state and players choose and play an action simultaneously : they receive rewards and then the state of the environment changes (possibly randomly), this depending on the current state and the joint choice of actions. At the early stage of the game, the reward function and transition probabilities are unknown by the learning agent. 

An important concept in stochastic games is the Nash equilibrium. In two-player zero-sum stochastic games, all Nash equilibria have the same value which represents a bound of the minimum expected average-reward that each player can obtain if they play optimally, regardless of the strategy of the opponent. When both agents play according to the Nash equilibrium, each of them can not hope a greater payoff by changing unilaterally his own strategy. Therefore, Nash equilibrium is often a benchmark for learning performances and a large set of works focus on this notion.

Two settings can be considered for learning in stochastic games: centralized and decentralized. In the first setting, a central learner tries to find the Nash equilibrium as quick as possible by controlling both players, \emph{e.g.} \cite{littman1994markov,sidford2020solving,bai2020provable,bai2020near,xie2020learning}.
In the second setting, the learner controls only one player and tries to be as efficient as possible against an arbitrary opponent, see \cite{brafman2002r,wei2017online,tian2021online,jafarnia2021learning,xie2020learning,sayin2021decentralized,liu2022learning,mao2022provably,sayin2022fictitious,wei2021last} for different settings and objectives. This setting is quite natural but more challenging than the first one because the learner has no control on the opponent strategy that may change over time. In this paper, we consider the decentralized learning problem. Several definitions for the regret are possible in this case, that we will discuss in more detail in Section 2.3. 

Learning algorithms are often split into two families called respectively model-based algorithms and model-free algorithms. In a model-based algorithm, the learner gathers observations of the game in order to estimate the parameters of the model (reward functions and transition probabilities), and then uses this estimation to compute a policy that is used for playing and gathering more information. In a model-free algorithm, the learner directly tries to estimate the values of the true model and plays optimally according to these value estimations. The two papers that are closest to ours are the algorithms UCSG in \cite{wei2017online} and PSRL-ZSG in \cite{jafarnia2021learning}. They both present a decentralized learning algorithm in average-reward SG and are both model-based. To the best of our knowledge, we propose in this paper the first model-free algorithm for the stochastic games with infinite-horizon average-reward objective. Model-free algorithms have the advantage that they often need considerably less memory space to work, and additionally they often enjoy much better time complexity, which motivates our work. They can also be generalized to other settings, such as linear SGs. The drawbacks of model-free is that, in the literature, model-based algorithms often achieve better performances than model-free. 

\subsection{Our contribution}

We propose an algorithm, that we call decentralized optimistic\linebreak Nash Q-learning (DONQ-learning). This algorithm is designed for infinite-horizon stochastic games with average-reward objective.  Our paper uses Q-learning optimism techniques similar to the one introduced by Jin \emph{et al.} in their seminal paper \cite{jin2018q} for finite-horizon MDPs.  This optimistic Q-learning has then been extended in two directions: (1) to infinite-horizon average-reward MDPs in \cite{wei2020model} by using an artificial discounted setting, and (2) to finite horizon stochastic games in \cite{bai2020near,tian2021online}. In this paper we combine these two lines of work to obtain an efficient learning algorithm with sublinear regret in infinite-horizon average-reward SGs. We show that the expected regret of DONQ-learning is sublinear in time and is bounded by $O(D(SAB\log(2T^2))^{\frac{1}{3}} T^{\frac{2}{3}})$, where $T$ is the time horizon, $S$ is the number of states, $A$ and $B$ are the size of learner's and of the opponent's actions sets, and $D$ a known upper bound of the span of the game. This result is similar to the bound obtained in \cite{wei2020model} for the regret in infinite-horizon MDP. In addition, our algorithm enjoys a low computational complexity and low memory space requirement compared to the previous works of \cite{wei2017online} and \cite{jafarnia2021learning} in the same setting. We also provide a high-probability bound for the regret of our algorithm.  Surprisingly, our analysis also shows a high probability upper bound for the regret of order $O(\sqrt{D \log(2/\delta)} T^{\frac{3}{4}})$, which has a dependency in $T$ that is significantly higher than for the expected regret. This higher dependency in $T$ is due to an additional term that appears because of our decentralized setting where the learner may play against a weak opponent. This additional term appears as a martingale difference sequence which disappears when the expected regret is considered. 

\paragraph*{Outline} The rest of the paper is organized as follows. We first introduce the model, the regret definition and some background on optimistic Q-learning in Section~\ref{sec:preliminaries}. We present our DONQ-learning algorithm in Section~\ref{sec:algorithm}. In Section~\ref{sec:theoretical}, we give the regret upper bounds (expected regret and regret with high probability), along with their proofs. We provide a proof of the most important technical lemmas in Section~\ref{sec:proofs}, and the proofs of some other lemmas are delegated to the Appendix. Finally, we conclude in Section~\ref{sec:conclusion}. 

\subsection{Related work}


{\bf Q-learning in MDP :} Q-learning algorithms appeared in 1989 in the seminal paper of Watkins \cite{watkins1989learning}. The first provably-efficient Q-learning appears in \cite{jin2018q}, where the authors add an optimism term to Q-learning and prove that the algorithm achieves low regret performances in finite-horizon MDPs. The authors use a smart choice of learning rates that we also use in this paper. The same techniques are then adapted in \cite{dong2019q} for the discounted setting to derive an algorithm with sample-complexity guaranties. The above two works are used in \cite{wei2020model} to derive an optimistic Q-learning for average-reward MDPs that has a regret bound of order $T^{\frac{2}{3}}$. This algorithm only assumes that the MDP is weakly-communicating. Our work is inspired from \cite{wei2020model} and extend their algorithm for the decentralized multi-agent setting.


{\bf Q-learning in SG :} For stochastic games, a Q-learning with optimism principle first appears in \cite{bai2020near} for finite-horizon problems. An adaptation to decentralized learning (still in finite-horizon SG) is given in \cite{tian2021online}. For the infinite-horizon setting, the authors of \cite{sayin2021decentralized} use Q-learning in discounted SG for decentralized learning but without optimism so they only establish convergence results. Our work differs from the above because we consider the infinite-horizon average-reward setting.

{\bf Other model-free approaches :} Some papers propose other model-free algorithms, different from Q-learning, that also use optimism techniques. This includes the following:\\
{\it V-learning :} An unavoidable assumption to use Q-learning like algorithms is that the learner must observe the opponent's action at each time. The V-learning algorithm is proposed in \cite{bai2020near} for centralized learning in finite-horizon SG, and then is adapted in \cite{tian2021online} to design an efficient algorithm for decentralized learning when the agents do not observe the action of their opponent. \\
{\it OOMD :} For infinite-horizon average-reward MDP, the authors of \cite{wei2020model} propose the OOMD algorithm that achieves a regret bound of order $\sqrt{T}$ when they assume the ergodicity assumption. They construct an adversarial multi-armed bandit instance for each state to design efficient policies. \\
{\it Gradient descent :} Gradient descent is another approach for decentralized learning in SG when the agent does not observe the action of its opponent. Some papers propose procedures based on the optimistic gradient descent to achieve Nash equilibrium identification guaranties after a finite time, for the finite-horizon setting in \cite{daskalakis2020independent}, and for the discounted setting in \cite{wei2021last}. 


{\bf Model-based algorithms :} To the best of our knowledge, all the algorithms that deal with learning in infinite-horizon average reward stochastic games are model-based. UCRL2 is introduced in \cite{auer2008near} and uses optimism to achieve sublinear regret bounds in infinite-horizon average-reward MDP with finite diameter. In \cite{wei2017online}, UCRL2 is adapted into UCSG for SG setting, and two possible definitions for the diameter are proposed. These definitions extend the notions of "communicating" MDP and "ergodic" MDP for SG. Under the weakest assumption, UCSG enjoys a regret bound of order $T^{\frac{2}{3}}$. In this paper we assume the same diameter assumption. In a recent work, inspired from PSRL algorithm for MDP, PSRL-ZSG algorithm is proposed in \cite{jafarnia2021learning} and achieves a Bayesian regret of order $\sqrt{T}$ under the same structure assumption. Our work does not improve the above regret bounds but the model-free property of our algorithm induces a lower computational complexity and memory space requirement than the previous model-based algorithms. Plus, we do not study Bayesian regret but worst-case expected regret or worst-case high probability regret.

\section{Preliminaries}
\label{sec:preliminaries}

\subsection{Stochastic game model and notations}

We consider two-player zero-sum simultaneous stochastic games (also known as Markov games in the literature), which generalize the standard Markov decision processes (MDPs) into the two-player setting where both players try to maximize their own payoff.  {\it Zero-sum} means that in this game, what a player wins corresponds exactly to what the other loses. We write the payoff for the point of the first player, that we call the {\it max-player}. This player tries to maximize the payoff whereas the second player, that we call the {\it min-player}, tries to minimize it. In this paper, the learner is the max-player and the opponent is the min-player. The term {\it simultaneous} means that the two players choose simultaneously how to interact with the environment at each time. Here we consider the infinite-horizon setting where players play indefinitely. At each time step, the game is in a current state known by the players and they play simultaneously an action. The players obtain a reward and the state of the game changes, both depending on the current state and the choices of actions. 

Formally, an instance of an infinite-horizon stochastic game is a tuple $(\mathcal{S}, \mathcal{A}, \mathcal{B}, r, P)$, where $\mathcal{S}$ is a set of states, $\mathcal{A}$ is the action set of the max-player, and $\mathcal{B}$ it the action set of the min-player. We assume that all the states and actions spaces are finite and of sizes $S$, $A$ and $B$.  For every $s, a, b \in \mathcal{S} \times \mathcal{A} \times \mathcal{B}$, $r(s, a, b) \in [0,1]$ is the reward that the max-player wins (and that the min-player loses) when the pair of actions $(a, b)$ is played in state $s$. Then the next state of the game is randomly drawn with the probability distribution $P(. | s, a, b)$. At each time step $t$ the players observe the states of the game $s_t$, they both choose simultaneously an action $a_t$ and $b_t$ that they play. They then observe the action of their opponent, receive the reward $r(s_t, a_t, b_t)$. The next state $s_{t+1}$ is then drawn according to $P(. | s_h, a_h, b_h)$.

{\bf Markovian policies :} A max-player Markovian policy $\mu : \mathcal{S} \rightarrow \Delta_{\mathcal{A}}$ is a function that maps every state $s$ with a probability distribution $\mu(.|s)$ on max-player's actions. Similarly, a min-player Markovian policy $\nu : \mathcal{S} \rightarrow \Delta_{\mathcal{B}}$ maps every state $s$ with a probability distribution $\nu(.|s)$ on min-player's actions. The pair of policies $(\mu, \nu)$ determines the rewards and the dynamics of the game. 
\medskip

{\bf Notations :} For a function $f : \mathcal{S} \rightarrow \R$, we write $Pf(s, a, b)$ for $\sum_{s'} P(s' | s, a, b)f(s')$. We also write $sp(f)$ for $\max_s f(s) - \min_s f(s)$.\\
For a function $g : \mathcal{S} \times \mathcal{A} \times \mathcal{B} \rightarrow \R$ and a pair of Markovian policies $(\mu, \nu)$, we write $\D_{(\mu, \nu)}[g](s) = \mathbb{E}_{a \sim \mu(.|s), b \sim \nu(. | s)}[g(s, a, b)]$. To lighten notation, we sometimes write $g^{(\mu, \nu)}(s)$ for this quantity. In particular, for the reward and transition probability, we have: 
\begin{align*}
    r^{(\mu, \nu)}(s) &= \mathbb{E}_{a \sim \mu(.|s), b \sim \nu(.|s)}[r(s, a, b)]\\
    P^{(\mu,\nu)}(s'|s) &= \mathbb{E}_{a \sim \mu(.|s), b \sim \nu(.|s)}[P(s'| s, a, b)].
\end{align*}

\subsection{Average-reward and game structure}

In this paper we establish low regret bounds for the undiscounted setting where players play indefinitely and try to maximize their average-reward over time. The average-reward of a pair of policies $(\mu,\nu)$ is defined for every state $s$, whenever the limit exists, by 
\begin{equation*}
    J^{(\mu,\nu)}(s) = \lim_{T \rightarrow +\infty} \frac{1}{T} \mathbb{E}_{(\mu,\nu)} \left[ \sum_{t = 1}^T r^{(\mu,\nu)}(s_t) \mid s_1 = s \right],
\end{equation*}
where $\mathbb{E}_{(\mu,\nu)}$ hides that the $(s_t)_t$ is a Markov chain whose transition kernel is $P^{(\mu,\nu)}$. This represents the expected average-reward for the max-player starting from state $s$ if both players play policies $\mu$ and $\nu$ respectively. 

Given a min-player policy $\nu$, a best response against $\nu$ for the undiscounted setting is a max-player policy $\mu^*$ such that for all state $s$, $J^{(\mu^*, \nu)}(s) = \max_{\mu} J^{(\mu, \nu)}(s)$. Similarly, given a max-player policy $\mu$, a best response against $\mu$ is a min-player policy $\nu^*$ such that for all state $s$, $J^{(\mu, \nu^*)}(s) = \min_{\nu} J^{(\mu, \nu)}(s)$.

To deal with the undiscounted setting, we assume the following structure property, which is known in the particular case of MDP, to be a minimal relevant framework for learning \cite{bartlett2012regal} :

\begin{asumption}\label{weakcomasump}
For every pair $(s, s')$ of states, and any min-player policy $\nu$, there exists a max-player policy $\mu$ such that $s'$ is accessible from $s$ with positive probability under policies $(\mu, \nu)$.
\end{asumption}
This assumption is intuitively necessary to get low regret because if this assumption does not hold, it means that the opponent may have a way to lock the learning agent in "bad stage" while at the early stage of the learning process, the agent could not have enough information to avoid this. Note that for the MDP setting, the analog of this assumption would be to assume that the MDP is "communicating": here the SG is communicating for the max-player's point of view, \emph{i.e.}, from the learner's point of view. In fact we may additionally assume the existence of a set of states that are transient under all Markovian policies, which would be the analogous property of "weakly communicating" for the MDP setting. This would not change our result but would complexify notations and proofs. 

As a consequence of Assumption~\ref{weakcomasump}, if $T^{(\mu,\nu)}_{s \rightarrow s'}$ is the expected time to go from $s$ to $s'$ under policies $(\mu, \nu)$, there exists a real number $D$ such that 
\begin{equation*}
    \max_{s, s'} \max_{\nu} \min_{\mu} T^{(\mu, \nu)}_{s \rightarrow s'} \leq D
\end{equation*}
The minimal $D$ is the diameter of the game \cite{wei2017online}. In this paper, we denote by $D$ an upper bound of the diameter and we suppose that this $D$ is known by the learning agent, so $D$ can be used to set some values of parameters.

We summarize important results that hold under Assumption~1 in the Theorem \ref{thframework} below. This theorem justifies the relevance of the learning objective in this framework. For a proof of these results, we refer to Theorem E.1 of \cite{wei2017online}.

\begin{theorem}[Theorem E.1 of \cite{wei2017online}]\label{thframework}
Under assumption \ref{weakcomasump}, there exist a unique real number $J^*$, a function $h^* : \mathcal{S} \longrightarrow \R$, and a pair of Markovian policies $(\mu^*, \nu^*)$ such that
\begin{enumerate}
    \item $J^* = \max_{\mu} \min_{\nu} J^{(\mu, \nu)}(s) = \min_{\nu} \max_{\mu} J^{(\mu, \nu)}(s)$ for every $s$ : the game value exists and does not depend on $s$.
    \item $(\mu^*, \nu^*)$ is a pair of Markovian policies such that $J^{(\mu^*, \nu^*)}(s) = J^*$ for all $s$ : $\mu^*$ is a best response against $\nu^*$ and vice-versa (Nash equilibrium).
    \item $sp(h^*) \leq D$ and satisfies the following Bellman's optimality equations :
    \begin{eqnarray*}
        J^* + h^*(s) & = & \max_{\mu} \left( r^{(\mu, \nu^*)}(s) + P^{(\mu, \nu^*)} h^*(s) \right), \\
        J^* + h^*(s) & = & \min_{\nu} \left( r^{(\mu^*, \nu)}(s) + P^{(\mu^*, \nu)} h^*(s) \right). 
    \end{eqnarray*}
\end{enumerate}
\end{theorem}

\subsection{Learning objective} 

In this paper, we consider a decentralized learning algorithm that has a low regret in a two-player zero-sum stochastic game. Nash equilibria in this setting provide to players an optimal way to play whatever their opponent does: If the max-player plays their Nash policy, it guarantees them an average-reward of at least $J^*$ against any opponent and we want to measure the online performance of the learning agent compared to this Nash policy. Thus our regret is given by 
\begin{equation}\label{defreg}
    \text{Reg}(T) = \sum_{t = 1}^T \left( J^* - r(s_t, a_t, b_t) \right) 
\end{equation}

This definition is classical for the regret in decentralized learning \cite{wei2017online,tian2021online,jafarnia2021learning}, and generalizes the definition for the MDP setting \cite{auer2008near,wei2020model}. Note that in contrast with MDPs, the regret is not necessarily non-negative: If the opponent is weak, then the learning agent can achieve an average-reward greater than $J^*$. The definition in Equation \eqref{defreg} ensures that, whatever the opponent's strategy,  the learner should learn a defensive strategy that works against it - but does not need to exploit the opponent's weakness - to achieve sublinear performances. 

A stronger regret definition would be to consider that the learner competes with the best response against the opponent (potentially assuming this one uses only Markovian policies). This would lead to the following regret definition:
\begin{equation*}
    \sum_{t = 1}^T \left( \max_{\mu} J^{(\mu, \nu_t)} - r(s_t, a_t, b_t) \right),
\end{equation*}
or a weaker version where the learner competes with the best fixed (over time) max-player policy, that is
\begin{equation*}
    \max_{\mu} \sum_{t = 1}^T \left( J^{(\mu, \nu_t)} - r(s_t, a_t, b_t) \right).
\end{equation*}
Unfortunately, it is shown in \cite{liu2022learning} that for the finite-horizon setting, it is not possible to design an algorithm that has a sublinear regret with the two above learning objectives.  Hence, and because the infinite-horizon setting is more general than the finite-horizon one, in this paper we focus on the regret definition given by \eqref{defreg}.

\subsection{Background on discounted SG}

As we will see later, our DONQ-learning algorithm, uses an artificial discounted setting to run.  Hence, we recall here some facts and notations about the discounted setting.

In the discounted setting, what players get at time $t$ is amortized by a  factor $\gamma^{t-1}$ where $\gamma < 1$ is called the discount factor. Given a discount factor $\gamma$ and a pair of policies $(\mu,\nu)$, the value function for the discounted setting is defined for every state $s$ as 
\begin{equation*}
    V_{\gamma}^{(\mu,\nu)}(s) = \mathbb{E}_{(\mu,\nu)} \left[ \sum_{t = 1}^{+\infty} \gamma^{t-1} r^{(\mu,\nu)}(s_t) \mid s_1 = s \right],
\end{equation*}
where $\mathbb{E}_{(\mu,\nu)}$ hides that the $(s_t)_t$ is a Markov chain whose transition kernel is $P^{(\mu,\nu)}$. This represents the expected discounted cumulative reward for the max-player starting from state $s$ if both players play policies $\mu$ and $\nu$ respectively. Then, for every pair of actions $(a, b)$, the Q-value function in $(s, a, b)$ is defined as 
\begin{equation*}
    Q_{\gamma}^{(\mu,\nu)}(s, a, b) = r(s, a, b) + \gamma PV_{\gamma}^{(\mu,\nu)}(s, a, b).
\end{equation*}
This represents the expected discounted cumulative reward for the max-player starting from state $s$ where players first play $a$ and $b$ respectively, and then follow policies $\mu$ and $\nu$ respectively. Therefore, for every state $s$  we have the relation 
\begin{equation*}
    V_{\gamma}^{(\mu,\nu)}(s) = \mathbb{D}_{(\mu,\nu)}[Q_{\gamma}^{(\mu,\nu)}](s).
\end{equation*}

Given a min-player policy $\nu$, a best response against $\nu$ for the $\gamma$-discounted setting is a max-player policy $\mu^*$ such that for all state $s$, $V_{\gamma}^{(\mu^*, \nu)}(s) = \max_{\mu} V_{\gamma}^{(\mu, \nu)}(s)$. Similarly, given a max-player policy $\mu$, a best response against $\mu$ is a min-player policy $\nu^*$ such that for all state $s$, $V_{\gamma}^{(\mu, \nu^*)}(s) = \min_{\nu} V_{\gamma}^{(\mu, \nu)}(s)$.

It is known that there always exists a pair of policies $(\mu^{\gamma}, \nu^{\gamma})$ such that $\mu^{\gamma}$ is a best response against $\nu^{\gamma}$ and vice-verse for the $\gamma$-discounted setting. $(\mu^{\gamma}, \nu^{\gamma})$ is called a Nash equilibrium of the game. All Nash equilibria have the same value function, denoted by $V_{\gamma}^*$, which satisfies for every state $s$
\begin{equation*}
    V_{\gamma}^*(s) = \max_{\mu} \min_{\nu} V_{\gamma}^{(\mu, \nu)}(s) = \min_{\nu} \max_{\mu} V_{\gamma}^{(\mu, \nu)}(s).
\end{equation*}
We also denote by $Q_{\gamma}^*$ the unique Q-value function of the Nash equilibria. $V_{\gamma}^*$ and $Q_{\gamma}^*$ satisfy the following Bellman's optimality equations for all $(s, a, b)$ :
\begin{align*}
    Q_{\gamma}^*(s, a, b) & = r(s, a, b) + \gamma PV_{\gamma}^*(s, a, b), \\
    V_{\gamma}^*(s) & = \max_{\mu} \min_{\nu} \mathbb{D}_{(\mu, \nu)}[Q_{\gamma}^*](s).
\end{align*}

We recall (see e.g. \cite{wei2017online} Lemma E.2) that under Assumption 1, if $D$ is a bound on the diameter of the game as defined above, then for any $\gamma < 1$ we have 
\begin{equation}\label{eqspinfD}
    sp(V_{\gamma}^*) \leq D.
\end{equation}

\subsection{Learning rate notations and properties}

In our DONQ-learning algorithm, we use optimistic Q-learning updates to maintain estimators. At a given time $t$, such an algorithm observes the current state-action triplet $(s,a,b)$ and updates the $Q$-value of this triplet as:
\begin{equation*}
    Q_{t+1}(s,a,b) \gets (1 - \alpha_{\tau}) Q_t(s,a,b) + \alpha_{\tau} (\text{[new sample]} + \text{[bonus]}),
\end{equation*}
where $\tau$ is the number of visits of the triplet $(s,a,b)$ up to time $t$.

The quantity $\alpha_{\tau}$ is called the learning rate and has to be well-chosen to ensure good learning performances. In the optimistic Q-learning algorithm for $H$-finite-horizon MDPs, the authors of \cite{jin2018q} propose the learning rate $\alpha_{\tau} = \frac{H + 1}{H + \tau}$ and they show that such a learning rate has good properties summarized in Lemma~\ref{lempropalpha} below.  For every $0 \leq \tau \leq T$, they define $\alpha_{\tau}^0 = \prod_{j = 1}^{\tau} (1 - \alpha_j)$ and for all $1 \leq i \leq \tau$, $\alpha_{\tau}^i = \alpha_i \prod_{j = i + 1}^{\tau} (1 - \alpha_j)$. Note that if $\tau \geq 1$, $\alpha_{\tau}^0 = 0$ and that $\sum_{i = = 0}^{\tau} \alpha_{\tau}^i = 1$.  The sequence  $\alpha_{\tau}^i$ has the following properties.

\begin{lemma}[Lemma 4.1 of \cite{jin2018q}]\label{lempropalpha}
The sequence $\alpha_{\tau}^i$ satisfy : 
\begin{enumerate}
    \item $\frac{1}{\sqrt{\tau}} \leq \sum_{i=1}^{\tau} \frac{\alpha_{\tau}^i}{\sqrt{i}} \leq \frac{2}{\sqrt{\tau}}$ for every $\tau \geq 1$.
    \item $\sum_{t = 1}^{\tau} (\alpha_{\tau}^i)^2 \leq \frac{2H}{\tau}$ for every $\tau \geq 1$.
    \item $\sum_{\tau = i}^{\infty} \alpha_{\tau}^i \leq 1 + \frac{1}{H}$ for every $i \geq 1$
\end{enumerate}
\end{lemma}

In this paper we use the same choice of learning rate for some carefully chosen parameter $H$.

\section{The DONQ-learning Algorithm}
\label{sec:algorithm}

The main idea of our decentralized optimistic Nash Q-learning (that we present in Algorithm~\ref{algo:DONQ}) is to learn the game in a well-chosen discounted setting using optimistic Q-Learning techniques first introduced in \cite{jin2018q}. The convenience of the $\gamma$-discounted setting is that the Q-values are well-defined and bounded by $1/(1 - \gamma)$ which enables to build and maintain optimistic estimators $\overline{Q}_t$ for the Nash Q-value $Q_{\gamma}^*$.

\begin{algorithm}
\caption{DONQ-learning Algorithm }
\label{algo:DONQ}
\begin{algorithmic}[1]
\State {\bf Parameters :} $H > 1$, $\delta \in (0, 1)$
\State {\bf Initialization :} $\forall s, a, b$
\State $\overline{Q}_1(s, a, b) = \overline{Q}^{\downarrow}_1(s, a, b) = \overline{V}^{\downarrow}_1(s) = H$
\State $N_1(s, a, b) = 0$ ; $\mu_1(a | s) = 1/A$ ; $\gamma = 1 - 1/H$
\State $\forall \tau \geq 1, \quad \alpha_{\tau} = \frac{H+1}{H+\tau} \quad \beta_{\tau} = 2 D \sqrt{\frac{H \log(2T/\delta)}{\tau}}$
\State {\bf Begin :}
\State Observe $s_1$.
\For{$t = 1, 2, \ldots, T$}
    \State Draw and play $a_t \sim \mu_t(. | s_t)$.
    \State Observe $b_t$ (drawn from the unknown opponent's policy).
    \State Observe $r(s_t, a_t, b_t)$ and $s_{t+1}$.
    \State Increment $N_{t+1}(s_t, a_t, b_t) \gets N_t(s_t, a_t, b_t) + 1$
    \State Set $\tau := N_{t+1}(s_t, a_t, b_t)$ and update $Q$-estimators as:
    \begin{align*}
        \overline{Q}_{t+1}(s_t, a_t, b_t) & \gets (1 - \alpha_{\tau}) \overline{Q}_t(s_t, a_t, b_t) +{} \\
        & \phantom{{}\gets{}} \alpha_{\tau} \left[ r(s_t, a_t, b_t) + \gamma \overline{V}^{\downarrow}_t(s_{t+1}) + \beta_{\tau} \right]\\
        \overline{Q}^{\downarrow}_{t+1}(s_t, a_t, b_t) &\gets \min{\left( \overline{Q}^{\downarrow}_t(s_t, a_t, b_t), \overline{Q}_{t+1}(s_t, a_t, b_t) \right)}
    \end{align*}
    \State Update the policy
    \begin{equation*}
        (\mu_{t+1}(. | s_t), \tilde{\nu}_{t+1}(.|s_t)) \gets \text{GetNashPolicies} \left( \overline{Q}^{\downarrow}_{t+1}(s_t, ., .) \right) 
    \end{equation*}
    \State Update $V$-estimator
    \begin{equation*}
        \overline{V}_{t+1}^{\downarrow}(s_t) \gets \mathbb{D}_{(\mu_{t+1}, \tilde{\nu}_{t+1})}[ \overline{Q}_{t+1}^{\downarrow}](s_t) 
    \end{equation*}
    \State $N_{t+1}$, $\overline{Q}_{t+1}$, $\overline{Q}_{t+1}^{\downarrow}$, $\overline{V}_{t+1}^{\downarrow}$, $\mu_{t+1}$ are kept equal to their values at the previous time step, $N_{t}$, $\overline{Q}_{t}$, $\overline{Q}_{t}^{\downarrow}$, $\overline{V}_{t}^{\downarrow}$, $\mu_{t}$, for all other $s, a, b$.
\EndFor
\end{algorithmic}
\end{algorithm}

The DONQ-learning algorithm takes as input a parameter $H$ and defines a discount factor $\gamma = 1 - \frac{1}{H}$. The agent then operates as if it was in a discounted objective. The algorithm also takes as parameter a confidence level $\delta \in (0, 1)$  which is set either to obtain a high probability regret bound, or to obtain a sublinear expected regret.

All optimistic estimators are initialized with $H$. At each time $t$, the learner plays an action $a_t$ drawn from the current policy $\mu_t$ and observes the opponent action $b_t$, the reward $r(s_t, a_t, b_t)$ and the next state $s_{t+1}$.  Then the optimistic estimator $\overline{Q}_{t+1}$ is updated for the current state-actions tuple $(s_t, a_t, b_t)$, by using an optimistic Q-learning like operation with a learning rate $\alpha_{\tau} = \frac{H+1}{H+\tau}$, the current optimistic estimator of the next state value $\overline{V}_t^{\downarrow}(s_{t+1})$ and a bonus term $\beta_{\tau}$ that scales in $D\sqrt{\frac{H}{\tau}}$, where $\tau$ is the number of visits of state-actions tuple $(s_t, a_t, b_t)$. For technical reasons, the algorithm maintains a non-increasing version of $\overline{Q}_t$, denoted by $\overline{Q}_t^{\downarrow}$, which is the one used to update the learner's policy and to compute the optimistic value estimators $\overline{V}_t^{\downarrow}$. 

The learner's policy is updated for the current state $s_t$ : $\mu_{t+1}(. | s_t)$ is chosen to be the max-player Nash policy of the matrix game $\overline{Q}_{t+1}^{\downarrow}(s_t, ., .)$. Note that if $\mu^*$ is an optimal Markovian Nash policy of the max-player, then for all $s$, $\mu^*(. | s)$ is a max-player Nash policy of the matrix game $Q^*(s, ., .)$. Then the algorithm also computes the min-player Nash policy $\tilde{\nu}_{t+1}(.|s_t)$ of the matrix game $\overline{Q}_{t+1}^{\downarrow}(s_t, ., .)$ in order to update $\overline{V}_{t+1}^{\downarrow}(s_t)$ as the expected value of $\overline{Q}_{t+1}^{\downarrow}(s_t, ., .)$ under policies $\mu_{t+1}$ and $\tilde{\nu}_{t+1}$. We emphasize that $\tilde{\nu}_{t+1}$ is not the policy used by the opponent at time $t+1$ (the opponent's policy is unknown to us and can be arbitrary).

\section{Theoretical guarantees}
\label{sec:theoretical}

\subsection{Main results: Regret Guarantees}

The key challenge is to correctly set the parameter $H$ to obtain sublinear regret guarantees. On the one hand, a large value for $H$ implies that the undiscounted objective is well approximated by the discounted case, but at the same time, the regret coming from the discounted analysis will be large. On the other hand, a small value of $H$ means that the undiscounted objective is not well approximated by the discounted case, which may have bad consequences the regret.

We give here two regret bounds, with high probability and in expectation, both with a sublinear dependency in $T$ obtained with a smart choice of the parameter $H$.  First, Theorem \ref{thhighprob} provides the high probability regret bound, depending on the parameter $H$.

\begin{theorem}\label{thhighprob}
For any $\delta \in (0, 1)$, there exists an absolute\footnote{By absolute, we mean a constant that does not depend on any of the models parameters.} constant $C$ such that with probability at least $1 - 3\delta$, the regret is bounded by 
\begin{align*}
    \text{Reg}(T) \leq & D \frac{T}{H}
    + 2H \sqrt{2T \log(2/\delta)} \\
    & + 12D \sqrt{SABHT \log(2T/\delta)} \\
    & + C\left(SABH + D \sqrt{T \log(4/\delta)} \right).
\end{align*}
\end{theorem}

If the parameter $H$ scales in $T^{\frac{1}{4}}$, this theorem shows a high probability regret bound of order $T^{\frac{3}{4}}$. Indeed, such a value of $H$ balances the terms $\frac{T}{H}$ and $H\sqrt{T}$ that appear in our analysis. 

\begin{corol}\label{coro:high_proba}
For any $\delta \in (0,1)$, we have that with probability at least $1 - 3\delta$, by setting the parameter $H = \sqrt{\frac{D}{\log(2/\delta)^{\frac{1}{2}}}}T^{\frac{1}{4}}$, for $T$ sufficiently large the regret is upper bounded by  
\begin{equation*}
    \text{Reg}(T) \leq \sqrt{D \log(2/\delta)^{\frac{1}{2}}} T^{\frac{3}{4}} 
    + o\left(T^{\frac{3}{4}}\right).
\end{equation*}
\end{corol}

The second term of the regret bound of Theorem~\ref{thhighprob} has a $H\sqrt{T}$ factor. This term does not appear in the analysis of \cite{wei2020model} for the MDP case. It is due to the fact that we were not able to bound $(\overline{Q}_t^{\downarrow} - Q_{\gamma}^*)(s_t, a_t, b_t)$ with a better bound than $H$ (see Equation~${\bf (s4)}$ in the regret decomposition below). This difficulty appears because the learner plays against an opponent that may be possibly weak. More explanations are given in Appendix \ref{apdisczeta}. This term appears because we simply use Azuma-Hoeffding's inequality for our high probability regret bound. However, since this term appears as terms of a martingale difference sequence, their sum can be bounded more efficiently when we consider the expected regret, as we see below.

Theorem~\ref{thexp} provides upper bounds for the expected regret, also depending on the parameter $H$. This bound is similar to the regret bound of Theorem~\ref{thhighprob} with $\delta=1/T$ but is significantly smaller because the second term of the regret bound of Theorem~\ref{thhighprob} disappears. 
\begin{theorem}\label{thexp}
There exists an absolute constant $C$ such that 
\begin{align*}
    \mathbb{E}[\text{Reg}(T)] \leq & D \frac{T}{H} 
    + 12D \sqrt{SABHT \log(2T^2)} \\
    & + C\left(SABH + D \sqrt{T \log(4T)} \right) .
\end{align*}
\end{theorem}

By setting a parameter $H$ that scales in $T^{\frac{1}{3}}$, we can balance the terms $\frac{T}{H}$ and $\sqrt{HT}$ (due to regret bound for discounted setting) and obtain a regret bound in expectation of order $T^{\frac{2}{3}}$. The bound that we obtain for the expected regret is then significantly smaller than the high probability bound of Corollary~\ref{coro:high_proba} because Theorem~\ref{thexp} does not has the term in $H\sqrt{T}$ as Theorem~\ref{thhighprob}.

\begin{corol}\label{coro:expected}
For $T$ sufficiently large, by setting the parameter $H = \left( \frac{T}{SAB\log(2T^2)} \right)^{1/3}$, the expectation of the regret is bounded by 
\begin{equation*}
    \mathbb{E}[\text{Reg}(T)] \leq D(SAB\log(2T^2))^{\frac{1}{3}} T^{\frac{2}{3}} + o\left(T^{\frac{2}{3}}\right).
\end{equation*}
\end{corol}

We highlight the fact that in the algorithm and our analysis, we only use that the diameter $D$ is an upper bound of $sp(h^*)$ (Theorem~\ref{thframework}) and $sp(V^*)$ : any other known quantity that satisfies these two properties can replace $D$ in the parameter setting and the theoretical guarantees.

\subsection{Computational complexity}

An interesting property  of Q-learning like algorithms is that only the values of the current state and state-action tuple are updated at each step. This contrasts with model-based approaches \cite{wei2017online,jafarnia2021learning} where some policies have to be entirely computed in the inner loop of the algorithm,  or with value-iteration like algorithms \cite{bai2020provable,xie2020learning} where a sweep of the state space is needed at each step. The only non-constant time operation in DONQ-learning is the call to GetNashPolicies procedure, but it is known that we can compute the value and the Nash policy of a zero-sum matrix game of size $A \times B$ in a polynomial time via linear programming \cite{khachiyan1979polynomial,gay1986karmarkar}, and this does depend on the size of the state space.

\subsection{Proofs of Theorems~\ref{thhighprob} and \ref{thexp}}

To show Theorem \ref{thhighprob} and \ref{thexp}, we decompose the regret as follows:
\begin{align*}
    \text{Reg}(T)  = & \sum_{t = 1}^T (J^* - r(s_t, a_t, b_t)) \\
    = & \sum_{t = 1}^T (J^* - (1 - \gamma)V^*_{\gamma}(s_t)) & \bf{(s1)} \\
    & + \sum_{t = 1}^T (V^*_{\gamma}(s_t) - \mathbb{D}_{(\mu_t, \nu_t)}[Q^*_{\gamma}](s_t) - \zeta_t) & \bf{(s2)}  \\
    & + \sum_{t = 1}^T (\mathbb{D}_{(\mu_t, \nu_t)}[Q^*_{\gamma}](s_t) - r(s_t, a_t, b_t) - \gamma V^*_{\gamma}(s_t)) & \bf{(s3)} \\
    & + \sum_{t = 1}^T \zeta_t & {\bf (s4)} 
\end{align*}
where $\zeta_t = \mathbb{D}_{(\mu_t, \nu_t)}[\overline{Q}_t^{\downarrow} - Q^*_{\gamma}](s_t) - (\overline{Q}_t^{\downarrow} - Q^*_{\gamma})(s_t, a_t, b_t)$. Note that $\{\zeta_t\}_t$ is a martingale difference sequence, bounded by $2H$ in absolute value, with respect to the $\sigma$-algebra generated by all random variables up to $s_t$ since $a_t$ and $b_t$ are drawn with policies $\mu_t(.|s_t)$ and $\nu_t(.|s_t)$ respectively. We bound {\bf (s1)}, {\bf (s2)} and {\bf (s3)} separately, sometimes under high probability events, and we finally deal with {\bf (s4)} specifically for both types of regret bounds. 

\subsubsection{Analysis of the term~\textbf{(s1)}}

The term ${\bf (s1)}$ represents the gap we pay because we learn in an artificial discounted setting. Each term of ${\bf (s1)}$ tends to 0 when $\gamma$ is close to 1, that is when $H$ is big. Concretely we show the following Lemma \ref{lems1} :
\begin{lemma}\label{lems1}
For every $s \in \mathcal{S}$,
\begin{equation*}
    \left| J^* - (1 - \gamma)V^*_{\gamma}(s) \right| \leq (1 - \gamma) D = \frac{D}{H}.
\end{equation*}
\end{lemma}
When summing over $t$ in {\bf (s1)}, a factor $T$ appears but the factor $1/H$ enables to get a sublinear dependency in $T$ in the regret with a well-chosen parameter $H$.

\subsubsection{Analysis of the term~\textbf{(s2)}}

Then, we first bound {\bf (s2)}  under an event $\mathcal{E}_{\delta}$ defined below for any $\delta \in (0,1)$ and that ensures in particular that $\overline{Q}_t^{\downarrow}$ and $\overline{V}_t^{\downarrow}$ are really upper bounds of $Q_{\gamma}^*$ and $V_{\gamma}^*$ respectively.

\noindent \fbox{\parbox{\linewidth}{
$\mathcal{E}_{\delta}$ := For all $(s, a,b)$ and time $t$, the following inequalities hold
\begin{equation}\label{qvpositif}
    (\overline{Q}_t^{\downarrow} - Q_{\gamma}^*)(s, a, b) \geq 0,
    \quad \quad 
    (\overline{V}_t^{\downarrow} - V_{\gamma}^*)(s) \geq 0,
\end{equation}
\begin{align}\label{relqv}
    (\overline{Q}_{t+1}^{\downarrow} - Q_{\gamma}^*)(s, a, b) \leq &
    \alpha_{\tau}^0 H + 6 D\sqrt{\frac{H}{\tau} \log(2T/\delta)} \nonumber \\
    & \phantom{\alpha_{\tau}^0 H} + \sum_{i = 1}^{\tau} \gamma \alpha_{\tau}^i \left[ (\overline{V}_{t_i}^{\downarrow} - V_{\gamma}^*)(s_{t_i + 1}) \right]
\end{align}
where $\tau = N_{t + 1}(s, a, b)$ and $t_1, t_2, \ldots, t_{\tau}$ the times where $(s, a, b)$ was visited.
}}\\

\begin{lemma}\label{lemE}
For any $\delta \in (0, 1)$, $\mathcal{E}_{\delta}$ is true with probability at least $1 - \delta$.
\end{lemma}

A proof of Lemma \ref{lemE} is given in Appendix \ref{aplemE}. $\mathcal{E}_{\delta}$ holds with probability at least $1 - \delta$ and under $\mathcal{E}_{\delta}$, Lemma \ref{lems2} gives an upper bound for ${\bf (s2)}$ that scales in $\sqrt{HT}$. This is the main contribution of the regret due to performances of optimistic Q-learning in the discounted setting.
\begin{lemma}\label{lems2}
For any $\delta \in (0, 1)$, under $\mathcal{E}_{\delta}$, there exists an absolute constant $c$ such that
\begin{align*}
    \sum_{t = 1}^T (V_{\gamma}^*(s_t) - \mathbb{D}_{(\mu_t, \nu_t)}[Q_{\gamma}^*](s_t) - \zeta_t) 
    &\leq 12 D\sqrt{SABHT \log(2T/\delta)}\\ &\qquad + cSABH.
\end{align*}
\end{lemma}

\subsubsection{Analysis of the term~\textbf{(s3)}}

To bound ${\bf (s3)}$ we define an other event $\mathcal{E}_{\delta}'$ for any $\delta \in (0, 1)$ that ensures that Azuma-Hoeffding's inequalities used on functions $V_{\gamma}^*$ and $Q_{\gamma}^*$ hold. 

\noindent \fbox{\parbox{\linewidth}{
$\mathcal{E}_{\delta}'$ : The two following inequalities hold
\begin{equation}\label{Ep1}
    \left| \sum_{t = 1}^T (\mathbb{D}_{(\mu_t, \nu_t)}[Q_{\gamma}^*](s_t) - Q_{\gamma}^*(s_t, a_t, b_t)) \right| \leq (1 + \gamma D) \sqrt{2T \log(4/\delta)}
\end{equation}
\begin{equation}\label{Ep2}
    \left| \sum_{t = 1}^T (PV_{\gamma}^*(s_t, a_t, b_t) - V_{\gamma}^*(s_{t+1}) ) \right|  \leq D \sqrt{2 T \log(4/\delta)}
\end{equation}
}}

\begin{lemma}\label{lemEp}
$\mathcal{E}_{\delta}'$ holds with probability at least $1 - \delta$
\end{lemma}

A proof of Lemma \ref{lemEp} is given in Appendix \ref{aplemEp}. In fact, the terms in ${\bf (s3)}$ are almost terms of Bellman's equations so, since $\mathcal{E}_{\delta}'$ holds with probability at least $1 - \delta$, we bound ${\bf (s3)}$ with high probability as shown in Lemma \ref{lems3} that is proved in Appendix \ref{aplems3}.
\begin{lemma}\label{lems3}
For any $\delta \in (0, 1)$, under $\mathcal{E}_{\delta}'$, there exists an absolute constant $c'$ such that
\begin{equation*}
    \sum_{t = 1}^T (\mathbb{D}_{(\mu_t, \nu_t)}[Q_{\gamma}^*](s_t) - r(s_t, a_t, b_t) - \gamma V_{\gamma}^*(s_t)) 
    \leq c'D \sqrt{T \log(4/\delta)}.
\end{equation*}
\end{lemma}

\subsubsection{Analysis of the term~\textbf{(s4)} and conclusion}

We prove the high probability regret bound by compiling the previous high probability bounds on {\bf (s1)}, {\bf (s2)}, and {\bf (s3)} (Lemmas \ref{lems1}, \ref{lems2} and \ref{lems3}). Additionally we use Azuma-Hoeffding to bound $\sum_{t = 1}^T \zeta_t$ with high probability. This makes appear a term that scales in $H\sqrt{T}$. We balance this term and the term $\frac{T}{H}$ of {\bf (s1)} with a well-chosen parameter $H$ of order $T^{\frac{1}{4}}$ to obtain an upper bound of order $T^{\frac{3}{4}}$ for the regret. More details are given in Appendix \ref{apdetthhighprob}.

We also establish the regret bound in expectation using the results of previous sections (Lemmas \ref{lems1}, \ref{lems2} and \ref{lems3}). To deal with the remaining $\sum_{t = 1}^T \zeta_t$, we use the fact that $\{\zeta_t\}$ is a martingale difference sequence (bounded sequence with a conditional expectation of 0 according to its filtration) to show that its weight is negligible especially in the case where the previous high probability events hold. We then set $H$ smartly to balance the dominant term $\frac{T}{H}$ and $\sqrt{HT}$ with respect to $T$, and obtain the desired regret bound in expectation of order $T^{\frac{2}{3}}$. More details are given in the Appendix~\ref{apdetthexp}.

\section{Proofs of main Lemmas}
\label{sec:proofs}

Here we give the proofs of Lemma~\ref{lems1} and~\ref{lems2}. All the other omited proofs are given in Appendix. Lemma~\ref{lems1} (resp. Lemma \ref{lems2}) induces the term that scales in $\frac{T}{H}$ (resp. $\sqrt{HT}$) in the statements of Theorems~\ref{thhighprob} and \ref{thexp}.  

\subsection{Proof of Lemma~\ref{lems1}}

We denote by $(\mu^*, \nu^*)$ and $h^*$ the Nash equilibrium for the undiscounted setting and the bounded function, both given by Theorem \ref{thframework}. We also denote by $(\mu^{\gamma}, \nu^{\gamma})$ a Nash equilibrium for the discounted setting with discount $\gamma$. 

By sub-optimality of $\mu^*$ against $\nu^{\gamma}$ for the discounted setting (first inequality), and the sub-optimality of $\nu^{\gamma}$ against $\mu^*$ for the undiscounted setting (second inequality), we get
\begin{align*}
    V^*_{\gamma}(s)  & \geq  V^{(\mu^*, \nu^{\gamma})}_{\gamma}(s) \\
    & = \mathbb{E}_{\mu^*, \nu^{\gamma}} \left[ \sum_{t = 1}^{+\infty} \gamma^{t - 1} r^{(\mu^*, \nu^{\gamma})}(\mathscr{s}_t) \quad | \mathscr{s}_1 = s \right] \\
    & \geq \mathbb{E}_{\mu^*, \nu^{\gamma}} \left[ \sum_{t = 1}^{+\infty} \gamma^{t - 1} \left( J^* + h^*(\mathscr{s}_t) - P^{(\mu^*, \nu^{\gamma})}h^*(\mathscr{s}_t) \right) \quad | \mathscr{s}_1 = s \right].
\end{align*}

Since $\mathscr{s}_{t + 1}$ is generated with transition probability $P^{(\mu^*, \nu^{\gamma})}(. | \mathscr{s}_t)$ in the stochastic process under $\mathbb{E}_{\mu^*, \nu^{\gamma}}$, the last term in the expectation can be seen as $h^*(\mathscr{s}_{t+1})$. Therefore we get 
\begin{align*}
    V^*_{\gamma}(s)  & \geq \frac{J^*}{1 - \gamma} + \mathbb{E}_{\mu^*, \nu^{\gamma}} \left[ \sum_{t = 1}^{+ \infty} \gamma^{t - 1} (h^*(\mathscr{s}_t) - h^*(\mathscr{s}_{t + 1})) \quad | \mathscr{s}_1 = s\right] \\
    & = \frac{J^*}{1 - \gamma} + h^*(\mathscr{s}_1) - \mathbb{E}_{\mu^*, \nu^{\gamma}} \left[ \sum_{t = 2}^{+ \infty} (\gamma^{t - 2} - \gamma^{t - 1})h^*(\mathscr{s}_t) \quad | \mathscr{s}_1  = s\right] \\
    & \geq \frac{J^*}{1 - \gamma} + \min_s h^*(s) - \max_s h^*(s) \sum_{t = 2}^{+\infty} (\gamma^{t - 2} - \gamma^{t - 1}) \\
    & \geq \frac{J^*}{1 - \gamma} - sp(h^*)  \\
    & \geq \frac{J^*}{1 - \gamma} - D.
\end{align*}

By the sub-optimality of $\nu^*$ against $\mu^{\gamma}$ in the discounted setting (first inequality), and the sub-optimality of $\mu^{\gamma}$ against $\nu^*$ in the undiscounted setting (second inequality), we have
\begin{align*}
V^*_{\gamma}(s) & \leq V^{(\mu^{\gamma}, \nu^*)}_{\gamma}(s) \\
& = \mathbb{E}_{(\mu^{\gamma}, \nu^*)} \left[ \sum_{t = 1}^{+\infty} \gamma^{t-1} r^{(\mu^{\gamma}, \nu^*)}(\mathscr{s}'_t) \quad | \mathscr{s}'_1 = s \right] \\
& \leq \mathbb{E}_{(\mu^{\gamma}, \nu^*)} \left[ \sum_{t = 1}^{+\infty} \gamma^{t-1} (J^* + h^*(\mathscr{s}'_t) - P^{(\mu^{\gamma}, \nu^*)}h^*(\mathscr{s}'_t)) \quad | \mathscr{s}'_1 = s \right].
\end{align*}

Here, $\mathscr{s}'_{t + 1}$ is generated with transition probability $P^{(\mu^{\gamma}, \nu^*)}(. | \mathscr{s}'_t)$ in the stochastic process under $\mathbb{E}_{(\mu^{\gamma}, \nu^*)}$, the last term in the expectation can be seen as $h^*(\mathscr{s}'_{t+1})$. Therefore we get 
\begin{align*}
    V^*_{\gamma}(s)  & \leq \frac{J^*}{1 - \gamma} + \mathbb{E}_{\mu^{\gamma}, \nu^*} \left[ \sum_{t = 1}^{+ \infty} \gamma^{t - 1} (h^*(\mathscr{s}'_t) - h^*(\mathscr{s}'_{t + 1})) \quad | \mathscr{s}'_1 = s \right] \\
    & = \frac{J^*}{1 - \gamma} + h^*(\mathscr{s}'_1) - \mathbb{E}_{\mu^{\gamma}, \nu^*} \left[ \sum_{t = 2}^{+ \infty} (\gamma^{t - 2} - \gamma^{t - 1})h^*(\mathscr{s}'_t) \quad | \mathscr{s}'_1 = s \right] \\
    & \leq \frac{J^*}{1 - \gamma} + \max_s h^*(s) - \min_s h^*(s) \sum_{t = 2}^{+\infty} (\gamma^{t - 2} - \gamma^{t - 1}) \\
    & \leq \frac{J^*}{1 - \gamma} + sp(h^*) \\
    & \leq \frac{J^*}{1 - \gamma} + D.
\end{align*}

Putting both parts together and multiplying by $(1 - \gamma)$ we obtain the desired statement.


\subsection{Proof of Lemma \ref{lems2}}

We simply write $V^*$ and $Q^*$ for $V^*_{\gamma}$ and $Q^*_{\gamma}$ respectively to lighten notations. 

For every time $t$ we have
\begin{equation*}
    V^*(s_t) - \mathbb{D}_{(\mu_t, \nu_t)}[Q^*](s_t) = (V^* - \overline{V}_t^{\downarrow})(s_t) + (\overline{V}_t^{\downarrow} - \mathbb{D}_{(\mu_t, \nu_t)}[Q^*])(s_t) .
\end{equation*}
The second difference is upper bounded by 
\begin{align*}
    \overline{V}_t^{\downarrow}(s_t) - \mathbb{D}_{(\mu_t, \nu_t)}[Q^*](s_t) & = \mathbb{D}_{\mu_t, \tilde{\nu}_t}[\overline{Q}_t^{\downarrow}](s_t) - \mathbb{D}_{(\mu_t, \nu_t)}[Q^*](s_t) \\
    & \leq \mathbb{D}_{(\mu_t, \nu_t)}[\overline{Q}_t^{\downarrow} - Q^*](s_t) \\
    & = (\overline{Q}_t^{\downarrow} - Q^*)(s_t, a_t, b_t) + \zeta_t \\
    & = (\overline{Q}_{t+1}^{\downarrow} - Q^*)(s_t, a_t, b_t) \\
    & \phantom{=}\hspace{.1cm} + (\overline{Q}_t^{\downarrow} - \overline{Q}_{t+1}^{\downarrow})(s_t, a_t, b_t) + \zeta_t .
\end{align*}
The inequality holds by definition of $\tilde{\nu}_t$ as a best-response against $\mu_t$ for the matrix game $\overline{Q}_t^{\downarrow}(s_t, ., .)$.
Under $\mathcal{E}_{\delta}$, inequality \eqref{relqv} bounds the term $(\overline{Q}_{t+1}^{\downarrow} - Q^*)(s_t, a_t, b_t)$, and therefore, 
\begin{align*}
    V^*(s_t) -  \mathbb{D}_{(\mu_t, \nu_t)}&[Q^*](s_t) - \zeta_t 
    \leq (V^* - \overline{V}_t^{\downarrow})(s_t)  \\
    & + (\overline{Q}_t^{\downarrow} - \overline{Q}_{t+1}^{\downarrow})(s_t, a_t, b_t) \\
    & + 6 D \sqrt{\frac{H}{n_t} \log(2T/\delta)} \nonumber \\
    & + \sum_{i = 1}^{n_t} \gamma \alpha_{n_t}^i \left[ (\overline{V}_{t_i(s_t, a_t, b_t)}^{\downarrow} - V^*)(s_{t_i(s_t, a_t, b_t) + 1}) \right],
\end{align*}
where $n_t = N_{t+1}(s_t, a_t, b_t)$ and $t_i(s_t, a_t, b_t)$ is the $i$-th time when $(s_t, a_t, b_t)$ has been  visited. Note that $\alpha_{n_t}^0 H = 0$ since\\ $n_t = N_{t+1}(s_t, a_t, b_t) > 0$ by definition. 

Now we sum over $t$ to get {\bf (s2)}. We need this technical Lemma :
\begin{lemma}\label{lemordersum}
Under $\mathcal{E}_{\delta}$, we have 
\begin{equation*}
     \gamma \sum_{t = 1}^T \sum_{i = 1}^{n_t} \alpha_{n_t}^i (\overline{V}_ {t_i(s_t, a_t, b_t)}^{\downarrow} - V^*)(s_{t_i(s_t, a_t, b_t) + 1})
    \leq SH + \sum_{t = 2}^{T+1} (\overline{V}_t^{\downarrow} - V^*)(s_t)
\end{equation*}
where $n_t = N_{t+1}(s_t, a_t, b_t)$ and $t_i(s_t, a_t, b_t)$ is the $i$-th time when $(s_t, a_t, b_t)$ has been visited.
\end{lemma}
A proof of Lemma \ref{lemordersum} is given in Appendix \ref{aplemordersum}.

\begin{itemize}
    \item By Lem.\ref{lemordersum}, $\sum_{t = 1}^T \sum_{i = 1}^{n_t} \gamma \alpha_{n_t}^i(\overline{V}_{t_i(s_t, a_t, b_t)}^{\downarrow} - V^*)(s_{t_i(s_t, a_t, b_t) + 1}) \leq SH + \sum_{t = 2}^{T+1} (\overline{V}_t^{\downarrow} - V^*)(s_t)$. This gives a telescoping sum with $\sum_{t = 1}^T (V^* - \overline{V}_t^{\downarrow})(s_t)$ and only $(\overline{V}_{T+1}^{\downarrow} - V^*)(s_{T+1}) - (\overline{V}_1^{\downarrow} - V^*)(s_1) \leq 2H$ remains. 

    \item Since $\overline{Q}_t^{\downarrow}$ decreases of at most $H$ per tuple $(s, a, b)$, we have $\sum_{t = 1}^T (\overline{Q}_t^{\downarrow} - \overline{Q}_{t+1}^{\downarrow})(s_t, a_t, b_t) \leq SABH.$

    \item Since $\sum_{s, a, b} N_{T + 1}(s, a, b) = T$, we have
    \begin{align*}
    & \sum_{t = 1}^T \frac{1}{\sqrt{n_t}} 
    = \sum_{t = 1}^T \sum_{s, a, b} \frac{{\bf 1}(s_t = s, a_t = a, b_t = b)}{\sqrt{N_{t+1}(s, a, b)}}
    = \sum_{s, a, b} \sum_{j = 1}^{N_{T + 1}(s, a, b)} \frac{1}{\sqrt{j}} \\
    & \leq \sum_{s, a, b} 2 \sqrt{N_{T+1}(s, a, b)}
    \leq 2 \sqrt{SAB \sum_{s, a, b} N_{T + 1}(s, a, b)}
    \leq 2 \sqrt{SABT}.
    \end{align*}
\end{itemize}

Putting every thing together, we get the desired inequality for a certain absolute constant $c$.


\section{Conclusion}
\label{sec:conclusion}

In this work we propose the DONQ-learning algorithm that is the first model-free algorithm that deals with decentralized learning in infinite-horizon average-reward stochastic games. It uses an artificial discounted setting in order to use optimism techniques developed in previous works. This algorithm achieves sublinear regret performances both with high probability and in expectation. Surprisingly, our analysis gives a high probability upper bound of order $T^{\frac{3}{4}}$ whereas we show an upper bound of order $T^{\frac{2}{3}}$ for the expected regret. We are not sure if this difference is a proof artifact or an actual difference between the expected and high-probability regret. The high probability regret bound is obtained by choosing a parameter $H=T^{1/4}$ while the bound for the expected regret is obtained by choosing $H=T^{1/3}$. An open question is whether choosing $H=T^{1/3}$ leads to a high probability bound in $T^{2/3}$ or in $T^{5/6}$ which is the high probability bound given by our analysis. Another open question is to know if there exists a model-free algorithm that achieves the information-theoretical lower bound of order $T^{\frac{1}{2}}$ under our Assumption~\ref{weakcomasump}. For future work, the same setting (decentralized learning in infinite-horizon average-reward SG) where we additionally assume that the opponent's actions are unobservable may be studied. This has already been studied for the finite-horizon setting with for instance the V-learning algorithm. It would be interesting to see if such model-free algorithms can be adapted to the infinite-horizon setting.


\bibliographystyle{ACM-Reference-Format} 
\bibliography{biblio}

\newpage

\appendix
\onecolumn

\section{Appendix}

Recall the regret definition and our decomposition of it :
\begin{align*}
    \text{Reg}(T) & = \sum_{t = 1}^T (J^* - r(s_t, a_t, b_t)) \\
    & = \sum_{t = 1}^T (J^* - (1 - \gamma)V^*_{\gamma}(s_t)) & \bf{(s1)} \\
    & + \sum_{t = 1}^T (V^*_{\gamma}(s_t) - \mathbb{D}_{(\mu_t, \nu_t)}[Q^*_{\gamma}](s_t) - \zeta_t) & \bf{(s2)}  \\
    & + \sum_{t = 1}^T (\mathbb{D}_{(\mu_t, \nu_t)}[Q^*_{\gamma}](s_t) - r(s_t, a_t, b_t) - \gamma V^*_{\gamma}(s_t)) & \bf{(s3)} \\
    & + \sum_{t = 1}^T \zeta_t & {\bf (s4)} 
\end{align*}
where $\zeta_t = \mathbb{D}_{(\mu_t, \nu_t)}[\overline{Q}_t^{\downarrow} - Q^*_{\gamma}](s_t) - (\overline{Q}_t^{\downarrow} - Q^*_{\gamma})(s_t, a_t, b_t)$. 

In what follows, we simply write $V^*$ and $Q^*$ for $V^*_{\gamma}$ and $Q^*_{\gamma}$ respectively to lighten notations. 

\subsection{Proof of Lemma \ref{lemE}}\label{aplemE}

This proof is adapted from the proof of Lemma 12 of \cite{wei2020model}: the proof is done for MDPs in \cite{wei2020model} and we adapt it to stochastic games.

 We have the following expression for $\overline{Q}_{t+1}(s, a,b)$ :
\begin{equation*}
    \overline{Q}_{t+1}(s, a,b) = \alpha_{\tau}^0 H + \sum_{i = 1}^{\tau} \alpha_{\tau}^i \left[ r(s, a, b) + \gamma \overline{V}_{t_i}^{\downarrow}(s_{t_i + 1}) + \beta_i \right].
\end{equation*}
By the Bellman equation on $Q^*$ and since $\sum_{i = 0}^{\tau} \alpha_{\tau}^i = 1$, this gives 
\begin{align*}
    (\overline{Q}_{t+1} - Q^*)(s, a,b) & = \alpha_{\tau}^0 (H - Q^*(s, a, b))
    + \sum_{i = 1}^{\tau} \alpha_{\tau}^i \left[ \gamma \overline{V}_{t_i}^{\downarrow}(s_{t_i + 1}) - \gamma PV^*(s, a, b) + \beta_i \right].
\end{align*}
By doing $\pm \gamma V^*(s_{t_i + 1})$ in the sum, and with $\tilde{\beta}_{\tau} = \sum_{i = 1}^{\tau} \alpha_{\tau}^i \beta_i$, we split the last into 
\begin{align}\label{splitq}
    (\overline{Q}_{t+1} - Q^*)(s, a,b) & = \alpha_{\tau}^0 (H - Q^*(s, a, b)) \nonumber \\
    &\quad + \sum_{i = 1}^{\tau} \gamma \alpha_{\tau}^i \left[ (\overline{V}_{t_i}^{\downarrow} - V^*)(s_{t_i + 1}) \right] \nonumber \\
    &\quad + \sum_{i = 1}^{\tau} \gamma \alpha_{\tau}^i\left[ V^*(s_{t_i + 1}) - PV^*(s, a, b) \right] \nonumber \\
    &\quad + \tilde{\beta}_{\tau} .
\end{align}

The first term is bounded between 0 and $\alpha_{\tau}^0 H $ because $Q^*(s, a, b)$ is bounded between 0 and $1/(1 - \gamma) = H$. 
The third term is a martingale difference sequence with each term bounded in \\ $[-\gamma \alpha_{\tau}^i sp(V^*), \gamma \alpha_{\tau}^i sp(V^*)]$, so by Azuma-Hoeffding's inequality, Lemma \ref{lempropalpha} and inequality \eqref{eqspinfD}, with probability at least $1 - \delta/T$, the third term is bounded in absolute value by $\gamma sp(V^*) \sqrt{2 \sum_{i = 1}^{\tau} (\alpha_{\tau}^i)^2 \log(2T/\delta)}) \leq 2 \gamma D \sqrt{\frac{H}{\tau} \log(2T/\delta)}$. Note that when $s, a, b, t$ vary, $\tau$ only takes its value in $1, 2, \ldots, T$ so with a union bound, the previous bound holds for all $(s, a, b, t)$ with probability $1 - \delta$. 
The forth term is bounded between $2 \gamma D \sqrt{\frac{H}{\tau}\log(2T/\delta)}$ and $4 \gamma D \sqrt{\frac{H}{\tau}\log(2T/\delta)}$ by Lemma \ref{lempropalpha}. 

By combining all aforementioned upper bounds (sometimes omitting $\gamma < 1$),  and since $\overline{Q}_{t+1}^{\downarrow} \leq \overline{Q}_{t+1}$, we have inequality \eqref{relqv}. 

We prove inequalities \eqref{qvpositif} by induction on $t$. Suppose that for a fixed time $t$, inequalities \eqref{qvpositif} hold for all $(s, a, b)$. By equation \eqref{splitq}, the induction hypothesis applied on the $(\overline{V}_{t_i}^{\downarrow} - V^*)(s_{t_i + 1})$, and aforementioned lower bounds, for all $(s, a, b)$ we have $(\overline{Q}_{t+1} - Q^*)(s, a, b) \geq 0$, so $(\overline{Q}_{t+1}^{\downarrow} - Q^*)(s, a, b) \geq 0$ by induction since $\overline{Q}_{t+1}^{\downarrow}(s, a, b) = \min(\overline{Q}_t^{\downarrow}(s, a, b), \overline{Q}_{t+1}(s, a, b))$. Then, 
\begin{align*}
    (\overline{V}_{t+1}^{\downarrow} - V^*)(s) & = \mathbb{D}_{\mu_{t+1}, \tilde{\nu}_{t+1}}[\overline{Q}_{t+1}^{\downarrow}](s) - \mathbb{D}_{\mu^{\gamma}, \nu^{\gamma}}[Q^*](s) \\
    & \geq \mathbb{D}_{\mu^{\gamma}, \tilde{\nu}_{t+1}}[\overline{Q}_{t+1}^{\downarrow}](s) - \mathbb{D}_{\mu^{\gamma}, \tilde{\nu}_{t+1}}[Q^*](s) \\
    & \geq 0,
\end{align*}
where $(\mu^{\gamma}, \nu^{\gamma})$ is the Nash equilibrium of $Q^*(s, ., .)$. The first inequality is by definition of $(\mu_{t+1}, \tilde{\nu}_{t+1})$ as a Nash equilibrium of $\overline{Q}_{t+1}^{\downarrow}(s, ., .)$.

\subsection{Proof of Lemma \ref{lemEp}}\label{aplemEp}

The event $\mathcal{E}_{\delta}'$ is the conjunction of two events that we can analyze separately:
\begin{itemize}
    \item The sequence $\mathbb{D}_{(\mu_t, \nu_t)}[Q^*](s_t) - Q^*(s_t, a_t, b_t)$ is a martingale difference sequence since $a_t$ and $b_t$ are drawn from the policies $\mu_t$ and $\nu_t$. By assumption of the span, each term is in an interval of length $1 + \gamma sp(V^*) \leq 1 + \gamma D$. Therefore, by Azuma-Hoeffding, with probability at least $1 - \frac{\delta}{2}$, Equation~\eqref{Ep1} holds.
    \item The sequence $PV^*(s_t, a_t, b_t) - V^*(s_{t+1})$ is a martingale difference sequence since $s_{t+1}$ is drawn according to $P(. | s_t, a_t, b_t)$. By assumption on the span, each term is in an interval of length $sp(V^*) \leq D$. Therefore by Azuma-Hoeffding, equation \eqref{Ep2} holds with probability at least $1 - \frac{\delta}{2}$. 
\end{itemize}

By the union bound, this shows that $\mathcal{E}_{\delta}'$ holds with probability at least $1 - \delta$.

\subsection{Proof of Lemma \ref{lems3}}\label{aplems3}

We split each term of the left-hand side of the main equation of Lemma~\ref{lems3} as follows:
\begin{align*}
    \mathbb{D}_{(\mu_t, \nu_t)}[Q^*](s_t) - r(s_t, a_t, b_t) - \gamma V^*(s_t)
    &= (\mathbb{D}_{(\mu_t, \nu_t)}[Q^*](s_t) - Q^*(s_t, a_t, b_t)) \\
    &\quad + (Q^*(s_t, a_t, b_t) - r(s_t, a_t, b_t) - \gamma PV^*(s_t, a_t, b_t) ) \\
    &\quad + \gamma (PV^*(s_t, a_t, b_t) - V^*(s_{t+1}) )\\
    &\quad + \gamma (V^*(s_{t+1}) - V^*(s_t) ).
\end{align*}

Now we sum each term :
\begin{itemize}
    \item Under $\mathcal{E}_{\delta}'$, $\sum_{t = 1}^T (\mathbb{D}_{(\mu_t, \nu_t)}[Q^*](s_t) - Q^*(s_t, a_t, b_t)) \leq (1 + \gamma D) \sqrt{2T \log(4/\delta)}$.
    \item The second term is 0 by Bellman's optimality equation for the discounted setting.
    \item Under $\mathcal{E}_{\delta}'$, $\gamma \sum_{t = 1}^T (PV^*(s_t, a_t, b_t) - V^*(s_{t+1})) \leq  D \sqrt{2T \log(4/\delta)}$.
    \item The last term gives a telescopic sum where $\gamma (V^*(s_{T+1}) - V^*(s_1)) \leq D$ remains by inequality \eqref{eqspinfD}.
\end{itemize}

Putting everything together finishes the proof.

\subsection{Proof of Lemma \ref{lemordersum}}\label{aplemordersum}

We simply adapt the proof of \cite{wei2020model} for the stochastic game setting. We split the sum of the left-hand side as follows to change the order of summation.
\begin{align*}
    \gamma \sum_{t = 1}^T \sum_{i = 1}^{n_t} \alpha_{n_t}^i [(\overline{V}_{t_i(s_t, a_t, b_t)}^{\downarrow} - V^*)(s_{t_i(s_t, a_t, b_t) + 1})]
    & = \gamma \sum_{t = 1}^T \sum_{s, a, b} {\bf 1}(s_t = s, a_t = a, b_t = b) \sum_{i = 1}^{N_{t+1}(s, a, b)} \alpha_{N_{t+1}(s, a, b)}^i [(\overline{V}_{t_i(s, a, b)}^{\downarrow} - V^*)(s_{t_i(s, a, b) + 1})] \\
   & = \gamma \sum_{s, a, b} \sum_{j = 1}^{N_{T+1}(s, a, b)} \sum_{i=1}^j \alpha_j^i [(\overline{V}_{t_i(s, a, b)}^{\downarrow} - V^*)(s_{t_i(s, a, b) + 1})] \\
   & = \gamma \sum_{s, a, b} \sum_{i = 1}^{N_{T+1}(s, a, b)} \sum_{j=i}^{N_{T+1}(s, a, b)} \alpha_j^i [(\overline{V}_{t_i(s, a, b)}^{\downarrow} - V^*)(s_{t_i(s, a, b) + 1})] \\
   & = \gamma \sum_{s, a, b} \sum_{i = 1}^{N_{T+1}(s, a, b)} [(\overline{V}_{t_i(s, a, b)}^{\downarrow} - V^*)(s_{t_i(s, a, b) + 1})] \left( \sum_{j=i}^{N_{T+1}(s, a, b)} \alpha_j^i \right)\\
   & \leq \gamma \sum_{s, a, b} \sum_{i = 1}^{N_{T+1}(s, a, b)} [(\overline{V}_{t_i(s, a, b)}^{\downarrow} - V^*)(s_{t_i(s, a, b) + 1})] \left( 1 + \frac{1}{H} \right), & \\
\end{align*}
where the last inequality is because by Lemma~\ref{lempropalpha} and the fact that $[(\overline{V}_{t_i(s, a, b)}]^{\downarrow} - V^*)(s_{t_i(s, a, b) + 1})] \geq 0$ under $\mathcal{E}_{\delta}$ (equation \eqref{qvpositif}).

As $\gamma \left(1 + \frac{1}{H} \right) \leq 1$, this is smaller than:
\begin{align*}
   & \leq \sum_{t = 1}^T [(\overline{V}_t^{\downarrow} - V^*)(s_{t+1})] \\
   & = \sum_{t = 2}^{T+1} [(\overline{V}_t^{\downarrow} - V^*)(s_t)] + \sum_{t = 1}^T [(\overline{V}_t^{\downarrow} - \overline{V}_{t+1}^{\downarrow}](s_{t+1}) \\
   & \leq \sum_{t = 2}^{T+1} [(\overline{V}_t^{\downarrow} - V^*)(s_t)] + SH,
\end{align*}
where the last inequality is because $\overline{V}_t^{\downarrow}$ decreases of at most $H$ per state $s$.

\subsection{Details for the proof of Theorem \ref{thhighprob}}\label{apdetthhighprob}

$\zeta_t$ is a martingale difference sequence which is bounded in $[-2H, 2H]$, so by Azuma-Hoeffding,
\begin{equation*}
    \left| \sum_{t = 1}^T \zeta_t \right| \leq 2H \sqrt{2T\log(2/\delta)}
\end{equation*}
with probability at least $1 - \delta$. By Lemma \ref{lemE} and Lemma \ref{lemEp}, $\mathcal{E}_{\delta}$, $\mathcal{E}_{\delta}'$ and this last inequality hold together with probability at least $1 - 3\delta$. Therefore, by Lemmas \ref{lems1}, \ref{lems2} and \ref{lems3}, with probability at least $1 - 3\delta$ the regret is upper bounded by 
\begin{align*}
    \text{Reg}(T) & \leq D \frac{T}{H} \\
    & + 2H \sqrt{2T \log(2/\delta)} \\
    & + 12D \sqrt{SABHT \log(2T/\delta)} \\
    & + C \left(SABH + D \sqrt{T \log(4/\delta)} \right),
\end{align*}
with $C = \max(c, c')$.

\subsection{Details for the proof of Theorem \ref{thexp}}\label{apdetthexp}

For any $\delta \in (0, 1)$, let $\mathcal{E}_{\delta}'' = \mathcal{E}_{\delta} \cap \mathcal{E}_{\delta}'$. By Lemma \ref{lemE} and Lemma \ref{lemEp}, $\mathcal{E}_{\delta}''$ holds with probability at least $1 - 2\delta$. 

The following inequality holds :
\begin{equation}\label{zetaind}
    \mathbb{E}\left[ {\bf 1}(\mathcal{E}_{\delta}'') \sum_{t = 1}^T \zeta_t \right] 
    \leq 4HT\delta .
\end{equation}
Indeed, we write 
\begin{equation*}
    \zeta_t =  {\bf 1}(\mathcal{E}_{\delta}'') \zeta_t + {\bf 1}(\neg \mathcal{E}_{\delta}'') \zeta_t,
\end{equation*}
where $\neg \mathcal{E}_{\delta}''$ is the complementary event of $\mathcal{E}_{\delta}''$. 
For any $t$, $\zeta_t$ is bounded by $2H$ in absolute value, therefore
\begin{equation*}
    - \mathbb{E} \left[ {\bf 1}(\neg \mathcal{E}_{\delta}'') \sum_{t = 1}^T \zeta_t \right] 
    \leq 2HT \mathbb{E} \left[ {\bf 1}(\neg \mathcal{E}_{\delta}'')  \right]
    \leq 4HT\delta.
\end{equation*}
Moreover, if $\mathcal{F}_t$ is the $\sigma$-algebra generated by all random variables up to $s_t$, since $\zeta_t$ is a martingale difference sequence for the filtration $\{ \mathcal{F}_t \}_t$, we have 
\begin{equation*}
    \mathbb{E} \left[ \sum_{t = 1}^T \zeta_t \right]
    = \mathbb{E} \left[ \sum_{t = 1}^T \mathbb{E} [ \zeta_t | \mathcal{F}_t ] \right] = 0
\end{equation*}
which proves inequality \eqref{zetaind}.

Finally the expectation of the regret is written as follows  for some $\delta \in (0, 1)$ that we smartly choose later :
\begin{equation}\label{regdelta}
    \mathbb{E}[\text{Reg}(T)] = \mathbb{E}[{\bf 1}(\mathcal{E}_{\delta}'') \text{Reg}(T)] + \mathbb{E}[{\bf 1}(\neg \mathcal{E}_{\delta}'') \text{Reg}(T)].
\end{equation}
By Lemma \ref{lems1}, Lemma \ref{lems2}, Lemma \ref{lems3}, and equation \eqref{zetaind}, the first term is upper bounded by  
\begin{align*}
    \mathbb{E} \left[ {\bf 1}(\mathcal{E}_{\delta}'') \text{Reg}(T) \right] 
    & \leq D \frac{T}{H} + 12 D \sqrt{SABHT \log(2T/\delta)} \nonumber \\
    & + \underbrace{ \mathbb{E} \left[{\bf 1}(\mathcal{E}_{\delta}'') \sum_{t = 1}^T \zeta_t \right]}_{\leq 4HT\delta} + C\left(SABH + D \sqrt{T \log(4/\delta)} \right),
\end{align*}
where $C = \max(c, c')$. 

Since the regret is clearly upper bounded by $T$, the second term of \eqref{regdelta} is upper bounded in expectation by 
\begin{equation*}
    \mathbb{E} \left[ {\bf 1}(\neg \mathcal{E}_{\delta}'') \text{Reg}(T) \right] 
    \leq 2T\delta .
\end{equation*}

By setting $\delta = \frac{1}{T}$, this gives the desired regret bound.

\subsection{Discussion for bounding $\zeta_t$}\label{apdisczeta}

The factor $H\sqrt{T}$ that appears in Theorem \ref{thhighprob} comes from the fact that we were not able to bound the terms in ${\bf (s4)}$ with a better bound than $H$. Indeed, consider a repeated version of Rock-Paper-Scissors ($S = 1$) and suppose the opponent always plays Rock. The optimistic estimations $\overline{Q}_t^{\downarrow}$ corresponding to the not-played opponent actions are never updated and remain equal to $H$, which leads to keep the estimator $\overline{V}_t^{\downarrow}$ equals to $H$. Therefore $\overline{Q}_t^{\downarrow}$ remains of order $H$ for all state-actions tuple whereas it is not the case for $Q^*$. Therefore it is not clear that the term $\zeta_t = (\overline{Q}_t^{\downarrow} - Q^*)(s_t, a_t, b_t) - \mathbb{D}_{(\mu_t, \nu_t)}[(\overline{Q}_t^{\downarrow} - Q^*)](s_t)$ can be efficiently bounded in order to use Azuma-Hoeffding's inequality on $\sum_{t=1}^T \zeta_t $.


\end{document}